\documentclass{article}
\usepackage{spconf,amsmath,amssymb,amsfonts,graphicx,hyperref}
\usepackage{textcomp,xcolor,booktabs,array,url}

\newcommand{\method}{SpectrumAudit}

\newcommand{\SubmissionJointRows}{%
PAMAP2 & FCN & 60.7 & 19.9$\pm$11.7 & 21.8$\pm$9.2 & 1.1$\pm$2.0 & -1.9$\pm$2.8 & 3/3 \\
PAMAP2 & ResCNN & 58.0 & 26.6$\pm$4.5 & 26.1$\pm$4.7 & 2.5$\pm$2.4 & 0.6$\pm$1.4 & 3/3 \\
PAMAP2 & Tr. & 64.3 & 21.7$\pm$8.2 & 23.2$\pm$9.1 & 4.7$\pm$4.1 & -1.5$\pm$1.1 & 3/3 \\
WISDM & FCN & 91.6 & 16.4$\pm$13.3 & 14.2$\pm$14.2 & 12.9$\pm$7.8 & 2.1$\pm$3.8 & 1/3 \\
WISDM & ResCNN & 91.0 & 23.2$\pm$16.2 & 21.1$\pm$17.4 & 14.8$\pm$0.4 & 2.1$\pm$3.9 & 1/3 \\
WISDM & Tr. & 85.3 & 12.1$\pm$5.3 & 15.0$\pm$9.3 & 9.1$\pm$5.1 & -3.0$\pm$13.4 & 1/3 \\
USC-HAD & FCN & 47.8 & 14.2$\pm$11.9 & 14.5$\pm$11.8 & 2.3$\pm$1.6 & -0.3$\pm$0.5 & 3/3 \\
USC-HAD & ResCNN & 46.8 & 17.4$\pm$6.3 & 19.8$\pm$3.4 & 5.9$\pm$1.2 & -2.4$\pm$3.9 & 3/3 \\
USC-HAD & Tr. & 49.4 & 27.0$\pm$8.3 & 17.0$\pm$4.1 & 1.8$\pm$2.7 & 10.0$\pm$11.8 & 3/3 \\
}

\newcommand{\SubmissionBaselineRows}{%
Random universal & none & -0.08$\pm$0.29 \\
Pseudo-label UAP--CE~\cite{rathore2020time} & target & 23.94$\pm$9.16 \\
Label-aware smooth UAP~\cite{pialla2022smooth} & target+Y & 15.99$\pm$8.45 \\
Auxiliary-ensemble transfer~\cite{huang2026transferability} & source & 21.46$\pm$9.93 \\
\textbf{\method{} (ours)} & target & 23.98$\pm$8.10 \\
}

\newcommand{\SubmissionFullRange}{2.87--40.83}
\newcommand{\SubmissionFullFOneRange}{5.71--35.86}
\newcommand{\SubmissionFullFOnePositive}{27}

\newcommand{\SubmissionVictims}{27}

\newcommand{\SubmissionConstantWins}{21}
\newcommand{\SubmissionConstantCellWins}{9}

\newcommand{\SubmissionOursBaselineMean}{23.98}
\newcommand{\SubmissionPseudoBaselineMean}{23.94}

\newcommand{\SubmissionPseudoComparisonP}{0.938}
\newcommand{\SubmissionMultiseedOursMean}{20.18}
\newcommand{\SubmissionMultiseedOursStd}{8.84}
\newcommand{\SubmissionMultiseedPseudoMean}{20.11}
\newcommand{\SubmissionMultiseedPseudoStd}{9.20}
\newcommand{\SubmissionMultiseedDifference}{+0.07}
\newcommand{\SubmissionMultiseedCILow}{-0.69}
\newcommand{\SubmissionMultiseedCIHigh}{0.72}
\newcommand{\SubmissionMultiseedP}{0.797}
\newcommand{\SubmissionMultiseedOursWins}{12}
\newcommand{\SubmissionMultiseedPseudoWins}{8}
\newcommand{\SubmissionMultiseedTies}{7}
\newcommand{\SubmissionPairedRepDCLarger}{24}
\newcommand{\SubmissionPairedRepDCNinety}{21}

\newcommand{\SubmissionInteractionP}{0.475}

\newcommand{\SubmissionProjectionACRescaled}{14}
\newcommand{\SubmissionProjectionDCLarger}{24}
\newcommand{\SubmissionProjectionDCNinety}{22}

\newcommand{\SubmissionProjectionDCRecoveryFailures}{5}

\newcommand{\SubmissionProjectionClusterDifference}{16.95}
\newcommand{\SubmissionProjectionClusterCILow}{10.86}
\newcommand{\SubmissionProjectionClusterCIHigh}{22.06}
\newcommand{\SubmissionProjectionClusterP}{0.0078}

\newcommand{\SubmissionCompositeMean}{10.26}
\newcommand{\SubmissionCompositeStd}{6.45}
\newcommand{\SubmissionCompositePositive}{7}
\newcommand{\SubmissionCompositeNegative}{2}

\title{WHEN TEMPORAL PERTURBATIONS ACT LIKE SENSOR BIASES:\
LABEL-FREE AUDITING OF WEARABLE ACTIVITY RECOGNIZERS}

\name{Qingyu Wu$^{1,*}$, Yuan Wei$^{1,*}$, Renju Liu$^{2}$, and Hua Cheng$^{1,\dagger}$}
\address{$^{1}$Defense Innovation Institute, Academy of Military Science, Beijing, China\\
$^{2}$School of Software Engineering, South China University of Technology, Guangzhou, China\\
$^{*}$Equal contribution; $^{\dagger}$Corresponding author; Email: \texttt{chenghua\_ams@163.com}}

\begin{document}
\ninept
\maketitle

\begin{abstract}
Wearable human-activity recognition (HAR) models operate across sensors,
subjects, and backbones, yet a smooth waveform may appear temporal
while exploiting primarily a persistent sensor offset. We introduce \method{},
a label-sealed audit that fits a phase-randomized full-window stimulus on
calibration windows from subjects held out from training and testing. After
selection, it replays its exact DC projection and budget-constrained zero-mean
residual on the same frozen victim without refitting. Across
\SubmissionVictims{} victims from three datasets and three backbones, the selected waveforms cause
\SubmissionFullRange{}-point three-phase robust accuracy losses. Under this
replay budget, DC is more damaging than AC on
\SubmissionProjectionDCLarger{}/27 victims and recovers at
least 90\% of the full drop on \SubmissionProjectionDCNinety{}/27; all
\SubmissionProjectionDCRecoveryFailures{} failures occur on WISDM. In a
held-out UTD-MHAD check, the selected waveform causes 13.49-pp
accuracy and 11.68-pp macro-F1 losses, versus $-0.66$ pp for matched random
changes. The audit diagnoses offset versus zero-mean variation under a common
peak-budget cap. The code will be released upon acceptance.
\end{abstract}


\section{Introduction}
Wearable human-activity recognition (HAR) underpins mobile health, assistive
technology, and context-aware systems. Its reliability depends on sensor
modality, subject composition, and deployment protocol~\cite{haresamudram2025tutorial}.
Recent self-supervised and lightweight architectures improve transfer and label
efficiency~\cite{logacjov2024ssl,crosshar2024,yuan2024persondays,
wavehar2025}. Sensor--language and wearable foundation models extend this line
of work to broader interfaces~\cite{qiu2025foundation,sensorlm2025}.
Yet a perturbation-induced failure may reflect the learned representation, input
normalization, or evaluation protocol rather than a temporal mechanism.

Adversarial perturbations offer controlled probes of these dependencies.
Prior work covers per-example and universal attacks, targeted objectives,
smooth parameterizations, and black-box search~\cite{fawaz2019adversarial,
rathore2020time,pialla2022smooth,pialla2025smooth,ding2023blackbox}.
In wearable and biosignal systems, attack behavior likewise depends on the
sensing modality and access model, spanning inertial wearables, Wi-Fi sensing,
and EEG systems~\cite{sah2019adar,xie2023universal,
kim2025realtime,zhong2025attention}. Attack success alone, however, does not
reveal which waveform component produces the failure. A smooth stimulus that
survives phase shifts may therefore appear temporal even when most of its effect
comes from a persistent offset.

\begin{figure}[t]
\centering
\includegraphics[width=0.98\columnwidth]{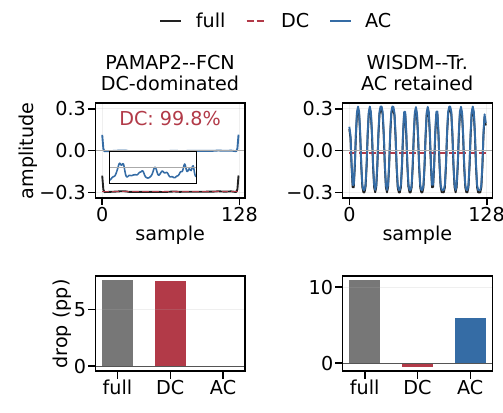}
\caption{Component-wise replay for smooth waveforms. (a) Illustrative
channels from selected PAMAP2 and WISDM waveforms, shown with their temporal means and
zero-mean residuals. The PAMAP2 trace is nearly flat (99.8\% DC energy); its
residual inset uses native amplitudes over samples 2--125. (b) Worst-phase
accuracy drops when the full waveform, its exact DC component, or its
budget-constrained AC residual is replayed on all channels. Both signals are
smooth, but their dominant components differ.}
\label{fig:gap}
\end{figure}

We ask whether a selected waveform acts through a persistent offset or time-varying
content. A separately optimized constant control can establish that an offset is
harmful, but cannot show that the selected full waveform derives its effect from
that offset.
Conversely, removing the temporal mean before optimization would preclude testing
whether that component matters. We separate selection from diagnosis: a waveform is
selected on unlabeled calibration windows from held-out subjects, then its exact
DC component and budget-constrained zero-mean residual are
replayed on the same frozen victim without refitting. Figure~\ref{fig:gap}
illustrates the distinction: signals with similar smooth appearance can have
different dominant components.

Our contributions are threefold. First, we establish a label-sealed,
subject-disjoint audit spanning three datasets, three backbone families, and
three independent victim seeds per setting. Second, we introduce a
select-then-project test for assessing whether a selected failure is carried
by a persistent offset or zero-mean temporal content. Third, we pair the diagnostic
with matched objective and access controls and exploratory victim-level analyses.
Together, these choices separate component-level diagnosis from both attack
strength and between-victim variation.

\section{Related Work}
Research on wearable HAR has expanded from task-specific convolutional and recurrent
classifiers to compact Transformers, self-supervised encoders, and cross-dataset learners
~\cite{zhou2022tinyhar,haresamudram2022assessing,logacjov2024ssl,
crosshar2024}. Large-scale pretraining and foundation encoders
further extend transfer across sensors and model families
~\cite{yuan2024persondays,wavehar2025,qiu2025foundation,
sensorlm2025}. This diversity makes a single-checkpoint evaluation inadequate.
We therefore span three task-trained backbone families. TimesFM is
designed for forecasting, whereas MOMENT is a general time-series encoder;
neither supplies native HAR logits without downstream adaptation
~\cite{das2024timesfm,goswami2024moment}. Evaluating either as a victim would
require a pre-specified HAR adapter and training policy; our claims do not
extend to adapted foundation models.

Time-series attacks include universal and per-example constructions, targeted and
untargeted objectives, smooth waveform parameterizations, and black-box search
~\cite{fawaz2019adversarial,rathore2020time,pialla2022smooth,pialla2025smooth,
ding2023blackbox}. Wearable and related biosignal settings add modality-specific
constraints, spanning inertial HAR, mmWave and Wi-Fi sensing, and
EEG~\cite{sah2019adar,xie2023universal,
kim2025realtime,zhong2025attention}. These studies are not directly commensurate
because victim access, supervision, perturbation budget, and phase protocol vary.
Accordingly, we adapt their objective and access principles to a common HAR
protocol and use the resulting controls to compare protocol conditions rather
than to rank published systems.

What remains unresolved is whether a selected waveform's temporal mean or
zero-mean variation carries its effect. Our label-sealed, subject-disjoint audit
addresses this question by replaying both components after selection on each
victim.

\section{Label-Sealed Audit}
\begin{figure*}[t]
\centering
\includegraphics[width=0.98\textwidth]{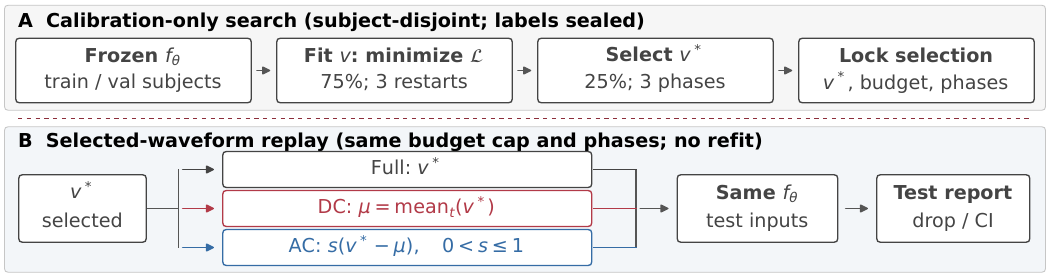}
\caption{Select-then-project protocol. (A) A frozen victim guides fitting on
75\% of unlabeled calibration windows and selection on the remaining 25\%
within the held-out calibration subjects.
(B) After selection is locked, three separate branches replay full, exact DC,
and zero-mean AC on the same victim; $s$ only reduces AC to the original
budget constraint. Test labels are used only for the final report.}
\label{fig:framework}
\end{figure*}

\textit{Overview.} Figure~\ref{fig:framework} summarizes the two-stage audit:
selection precedes attribution. A frozen victim guides waveform fitting and
selection on unlabeled calibration windows; the locked waveform is then replayed
on that victim in full and after DC/AC projection.

\subsection{Threat model and protocol}
\label{sec:protocol}
Let a frozen classifier return logits $\ell(x)$ and features $h(x)$ for a
standardized window $x\in\mathbb{R}^{C\times T}$. We consider an auditor that
receives unlabeled windows from held-out subjects; labels remain sealed for the
audit, pseudo-label, and transfer controls. We include label-aware smooth UAP
only as a privileged diagnostic; it is outside the admissible label-free audit.
We set $T=128$ and constrain every waveform by
$\lVert v\rVert_\infty\leq\epsilon=0.30$ in training-standard-deviation
units. Victim training uses a subject-disjoint validation split; the
calibration windows are split 75/25 for optimization and unlabeled selection;
this is a window-level holdout, and final test subjects remain separate.
Victims are trained with Adam (learning rate $10^{-3}$, weight decay
$10^{-4}$) for at most 40 epochs, with early stopping after six stale
validation epochs.
Per-channel raw-unit allowances are $0.10$--$7.32$ (PAMAP2), $1.46$--$2.11$
(WISDM), and $0.09$--$15.89$ (USC-HAD); differing units preclude perceptual
or physical realizability claims.

The audit assumes a frozen differentiable HAR classifier with activity logits
and an intermediate representation. We evaluate PAMAP2
~\cite{reiss2012pamap2}, WISDM~\cite{kwapisz2011wisdm}, and USC-HAD
~\cite{zhang2012uschad}, each paired with an FCN, a residual CNN, and a patch
Transformer. Each of the nine dataset--backbone settings has three independently
trained victim seeds, giving \SubmissionVictims{}
victims.
Three-phase robust drop is the minimum over shifts 0, 42, and 85, not all 128.
Mainline cells summarize three seeds; Table~\ref{tab:controls} uses one per cell.

\subsection{Probes and objectives}
For each victim, the full-window probe optimizes $v\in\mathbb{R}^{C\times T}$
with two sample-domain low-pass passes; their physical cutoffs therefore vary
with the dataset sampling rate. During fitting, each window receives a random circular
shift, $x_\phi=x+\operatorname{roll}_\phi(v)$. We retain DC during search so that
failures due to a persistent offset remain observable. After selection, we assess
component dominance by replaying the exact DC component of the frozen waveform
and budget-constrained zero-mean residual.
With clean pseudo-labels $\hat y=\arg\max_j\ell_j(x)$, the label-free objective is
\begin{equation}
\begin{aligned}
\mathcal{L}(v)={}&-\operatorname{CE}(\ell(x_\phi),\hat y)
-\lambda_h d_{\cos}(h(x),h(x_\phi))\\
&-\lambda_J\operatorname{JS}(p,p_\phi)
+\lambda_{TV}\operatorname{TV}(v),
\end{aligned}
\end{equation}
where $d_{\cos}$ is one minus cosine similarity; $p$ and $p_\phi$ denote
the clean and attacked softmax distributions. We set
$(\lambda_h,\lambda_J,\lambda_{TV})=(1,0.25,0.10)$.
The unlabeled selector maximizes $J_{\min}+0.25\bar J+0.10C_{\min}
+0.025\bar C+0.05\max(D_{\min},0)$ across checkpoints and restarts;
$J$, $C$, and $D$ denote phase-wise JS divergence, prediction-change rate,
and clean-pseudo-label probability drop (subscripts/overbars: phase
minimum/mean). The constant control learns one bounded value per channel;
zero-mean AC uses eight rate-aligned Fourier components at $0.5$--$8$ Hz.
Smoothness constrains the probe but does not establish a temporal mechanism.

After selection, we decompose each full-window waveform without refitting or
reselection. Its exact DC projection is the per-channel temporal mean,
$v_{\mathrm{DC}}=T^{-1}\sum_t v_t$, and its AC residual is
$v_{\mathrm{AC}}=v-v_{\mathrm{DC}}$. If cancellation in $v$ makes
$\lVert v_{\mathrm{AC}}\rVert_\infty>\epsilon$, only the residual is scaled
down to the original budget; its temporal mean remains zero. Replaying the
full signal, exact DC, and budget-constrained AC on the same frozen
victim provides a within-victim test of whether the selected stimulus relies on
persistent or time-varying content.
We predefine DC dominance as $\Delta_{\rm DC}>\Delta_{\rm AC}$ and
90\% recovery as $\Delta_{\rm DC}\geq0.9\Delta_{\rm full}$ under the worst
phase.

\subsection{Matched comparison controls}
All controls share the subject split, $\ell_\infty$ budget, and three-phase
test metric, while their waveform parameterizations and model access differ. Each
target-model full-window search uses three paired Adam restarts (learning
rate $0.03$, batch size 256, 600 updates) with checkpoints every 50 updates.
Pairing fixes probe initialization, minibatch order, and phase sequence;
the label-aware row also uses a label-based selection score. The
independently optimized constant and Fourier AC controls use their own
parameterizations: constant runs for 350 updates with restart-level selection;
AC runs for 600 updates with 50-step selection. Constant uses the audit
objective, whereas AC uses pseudo-label CE. Auxiliary-ensemble transfer fits
and selects on source models only; random universal is unfitted. The
literature-derived rows are protocol-matched adaptations, not reproductions.

\textbf{Random universal} uses no victim logits. \textbf{Pseudo-label UAP--CE}
removes the representation and JS terms, adapting Rathore et
al.~\cite{rathore2020time} to the common smooth parameterization
~\cite{pialla2022smooth}. \textbf{Label-aware smooth UAP} uses calibration
labels for fitting and selection, outside the admissible audit; this row does
not isolate label use during fitting alone.
\textbf{Auxiliary-ensemble transfer} trains on the two non-target
backbones, adapting Huang et al.~\cite{huang2026transferability}, and uses the
target only at evaluation. Test labels remain sealed until every
candidate is frozen. Table~\ref{tab:controls} contrasts these objective and
access choices under the common protocol.

\section{Evaluation}
\begin{table}[!b]
\centering
\small
\setlength{\tabcolsep}{3.0pt}
\caption{Matched controls on nine development settings
(one fixed victim per dataset--backbone cell; mean$\pm$SD across cells). Literature-derived rows
are protocol-matched adaptations, not reproductions of published rates. ``target'' means the frozen target
checkpoint is used during fitting; ``source'' means only the two non-target
backbones are used; ``target+Y'' uses calibration labels in fitting and
selection. Drops are in percentage points; Y denotes calibration labels.}
\label{tab:controls}
\begin{tabular*}{\columnwidth}{@{\extracolsep{\fill}}lcr@{}}
\toprule
Method & Access & Drop (pp)\\
\midrule
\SubmissionBaselineRows
\bottomrule
\end{tabular*}
\end{table}

\begin{table*}[t]
\centering
\small
\setlength{\tabcolsep}{3.3pt}
\caption{Mainline and independently optimized controls on 27 frozen victims
($\epsilon=0.30$); full-window, Const., AC, and $\Delta_{F-C}$ entries are
mean$\pm$sample SD over three seeds, whereas Clean is the three-seed mean;
accuracy drops are in percentage points. Const. and AC are independently optimized
mechanism controls, not matched attack rankings or post-selection projections.
$\Delta_{F-C}$ is full minus
constant in three-phase worst-case drop; C$>$AC counts constant wins over AC.}
\label{tab:main}
\begin{tabular*}{\textwidth}{@{\extracolsep{\fill}}llrrrrrr@{}}
\toprule
Dataset & Victim & Clean (\%) & Full-window & Const. ctrl. & AC ctrl. & $\Delta_{F-C}$ & C$>$AC\\
\midrule
\SubmissionJointRows
\bottomrule
\end{tabular*}
\end{table*}

We first isolate objective and access effects with paired target-model searches
and matched controls. We then characterize victim variation before attributing
selected failures to waveform components.

\subsection{Matched objective and access controls}
The first question is whether the representation and distribution terms change
attack strength under paired search trajectories. Table~\ref{tab:controls}
reports a nine-setting development grid with one fixed victim per
dataset--backbone cell; it is not a population estimate. Each paired search shares
the probe initialization, minibatch order, and phase sequence, while the
auxiliary-transfer row withholds the target checkpoint during fitting.

On this development grid, the representation-aware and pseudo-label objectives
yield mean drops of
\SubmissionOursBaselineMean{} and
\SubmissionPseudoBaselineMean{} pp, respectively, with a 3/3/3
ours/pseudo/tie split ($p=\SubmissionPseudoComparisonP$). The other rows provide
no-model, label-aware, and source-only reference points. Across 27 victims, the
paired search yields
\SubmissionMultiseedOursMean{}$\pm$
\SubmissionMultiseedOursStd{} pp for the proposed objective and
\SubmissionMultiseedPseudoMean{}$\pm$\SubmissionMultiseedPseudoStd{} pp for
UAP--CE (\SubmissionMultiseedOursWins{}/\SubmissionMultiseedPseudoWins{}/
\SubmissionMultiseedTies{} wins/losses/ties; cluster sign-flip
$p=\SubmissionMultiseedP$). The paired difference (ours minus UAP--CE) is
\SubmissionMultiseedDifference{} pp (95\% setting-cluster bootstrap CI
$[\SubmissionMultiseedCILow,\,\SubmissionMultiseedCIHigh]$). These results
do not establish an attack-strength advantage; the mechanism claim is instead
tested by post-selection component replay.

\subsection{Across-victim variation}

We next ask how the full-window effect varies across 27 victims before
attributing losses to waveform components. Table~\ref{tab:main} reports a separate
mainline search and independently optimized constant and zero-mean controls.
Across individual victims, the three-phase robust accuracy drop spans
\SubmissionFullRange{}
percentage points. Robust macro-F1 also decreases on all
\SubmissionFullFOnePositive{} victims, spanning \SubmissionFullFOneRange{}
percentage points, so the effect is not confined to accuracy. All 27 clean
checkpoints exceed the test-set majority-class baseline (13.2\%, 36.1\%, and
19.8\% for PAMAP2, WISDM, and USC-HAD), although USC-HAD mean clean accuracy
is 46.8--49.4\% across backbones. These data do not support an architecture
ranking: the Transformer drop changes from 27.0 pp on USC-HAD to 12.1 pp on
WISDM.

\begin{figure*}[t]
\centering
\includegraphics[width=0.98\textwidth]{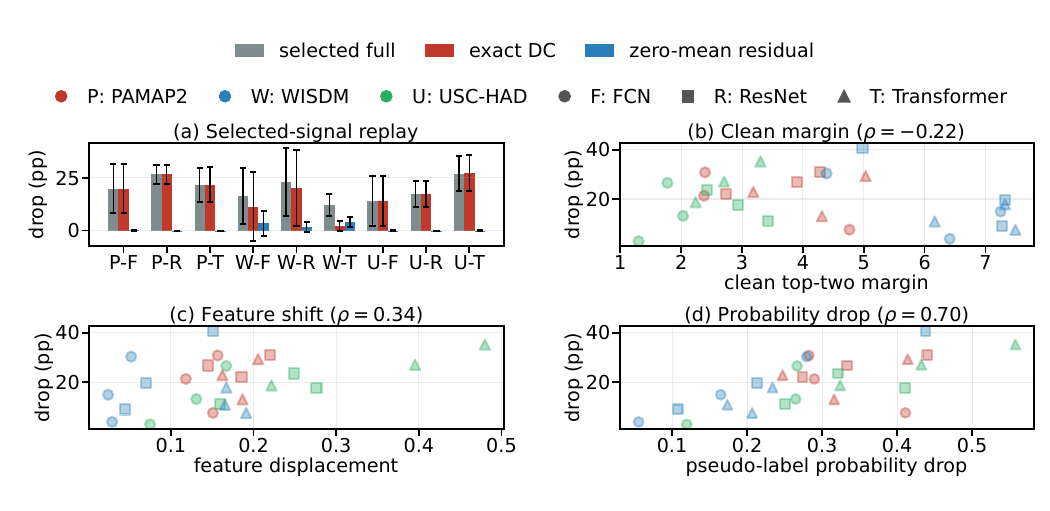}
\vspace{-9pt}
\caption{Component replay and exploratory diagnostics for 27 frozen victims.
(a) Replays of the same selected waveform: mean$\pm$sample SD over three
victim seeds per setting. DC exceeds AC on \SubmissionProjectionDCLarger{}/27
victims and recovers at least 90\% of the full drop on
\SubmissionProjectionDCNinety{}/27. AC is scaled down to budget on
\SubmissionProjectionACRescaled{}/27 victims.
(b)--(d) Each marker is one victim seed; colors and shapes encode the
dataset and backbone, respectively (see legend).
All drops are three-phase worst-case accuracy losses in percentage points.
$\rho$ is the unadjusted Spearman correlation, not a causal effect.}
\label{fig:mechanism}
\end{figure*}

Panels~\ref{fig:mechanism}(b)--(d) relate this variation to victim diagnostics.
Feature displacement shows a weaker unadjusted trend than pseudo-label
probability drop. Fixed-effect regressions estimate positive coefficients for
both post-intervention quantities (HC3 $p<0.004$), but not for clean margin or
input-gradient norm.
These analyses are exploratory and non-causal. The dataset--backbone interaction
is not significant ($p=\SubmissionInteractionP$); the cell differences do not
justify a claim of intrinsic Transformer robustness. We treat each victim seed as a replication
within its dataset--backbone cell and keep the conclusion at the level of audit
behavior.

As a check beyond the three-dataset benchmark, we repeat the label-sealed search
on one FCN victim trained on UTD-MHAD~\cite{chen2015utdmhad} and evaluate
held-out subjects 7--8.
The selected waveform causes a 13.49-pp three-phase robust accuracy drop
and an 11.68-pp macro-F1 drop,
whereas a matched random waveform changes accuracy by $-0.66$ pp. We treat this
single result as a three-phase protocol check rather than a population estimate.

\raggedbottom

\subsection{Component-level attribution}

We finally replay each selected waveform without further optimization to
determine which component retains the effect under the common perturbation
budget. Post-selection replays yield larger DC than AC drops on
\SubmissionProjectionDCLarger{}/27 victims (9/9 on PAMAP2 and USC-HAD;
6/9 on WISDM) and at least 90\% recovery on
\SubmissionProjectionDCNinety{}/27; all five recovery failures occur on WISDM,
including all three Transformer seeds.
Among the independently optimized controls, the constant control
exceeds AC on \SubmissionConstantWins{}/27 victims
and \SubmissionConstantCellWins{}/9 cell means.
The paired-objective replications show the same pattern: larger DC than AC
drops on \SubmissionPairedRepDCLarger{}/27 and at least 90\% recovery on
\SubmissionPairedRepDCNinety{}/27 for each objective.
The mainline AC residual requires budget scaling on
\SubmissionProjectionACRescaled{}/27 victims, so these are budget-constrained
replays, not an equal-energy comparison or an additive attribution of loss.
An equal-peak-budget DC+AC composite averages
\SubmissionCompositeMean{}$\pm$\SubmissionCompositeStd{} pp across phases.
Relative to the larger isolated component, its phase-averaged drop is higher on
\SubmissionCompositePositive{} settings and lower on
\SubmissionCompositeNegative{} settings.
Across the nine cell means, the DC--AC three-phase robust-drop difference is
\SubmissionProjectionClusterDifference{} pp (95\% setting-cluster bootstrap CI
$[\SubmissionProjectionClusterCILow{},\,\SubmissionProjectionClusterCIHigh{}]$;
cluster sign-flip $p=\SubmissionProjectionClusterP$). We report this
cell-weighted summary as a secondary check; the primary evidence remains the
victim-level replay counts.

As a phase-alignment check, we also replayed the 27 selected full waveforms
at all 128 circular shifts. Relative to the all-shift minimum, the three-phase
minimum is higher by 0.31 pp on average and 1.59 pp at most. The component
counts remain three-phase results: DC here denotes a constant component in
standardized model input, not an identified physical sensor bias.

\section{Conclusion}
Across the evaluated task-trained HAR victims, component replays indicate that
selected failures are usually DC-dominated rather than carried by zero-mean
variation under the tested phases and peak budget. The broader lesson is that a
smooth, phase-robust perturbation should not be interpreted as temporal evidence
without component-matched replay; the select-then-project audit provides a
label-sealed way to make that distinction. The present protocol targets
differentiable, task-trained victims and sample-domain smoothness, while physical
sensing, black-box transfer, and foundation-model backbones remain natural next
steps.

\bibliographystyle{IEEEbib}
\pagebreak
\section*{Compliance with Ethical Standards}
We use public wearable-sensor datasets; no new human or animal data. OpenAI
GPT-5.6 assisted with language polishing, parts of the experimental code, and
figure-generation/layout support; the authors take full responsibility for
independently checking text, citations, figures, code, permissions, and LLM-use
policy compliance.

\makeatletter
\renewcommand{\thebibliography}[1]{%
  \section{References}\list{[\arabic{enumi}]}{%
    \settowidth\labelwidth{[#1]}%
    \leftmargin\labelwidth\advance\leftmargin\labelsep
    \usecounter{enumi}%
    \itemsep -1.8pt plus 0pt minus 0.2pt
    \parsep 0pt}%
  \def\newblock{\hskip .11em plus .33em minus .07em}%
  \sloppy\clubpenalty4000\widowpenalty4000
  \sfcode`\.=1000\relax}
\makeatother
\bibliography{main}

\end{document}